\documentclass{article}

\usepackage{arxiv}
\usepackage[T1]{fontenc}    
\usepackage{hyperref}       
\usepackage{url}            
\usepackage{booktabs}       
\usepackage{amsfonts}       
\usepackage{nicefrac}       
\usepackage{microtype}      
\usepackage{lipsum}
\usepackage{graphicx}
\usepackage{amsmath,amssymb}

\usepackage{changepage}

\usepackage[utf8x]{inputenc}
\usepackage{textcomp,marvosym}

\usepackage{cite}

\usepackage{nameref,hyperref}

\usepackage[right]{lineno}

\usepackage{microtype}
\DisableLigatures[f]{encoding = *, family = * }

\usepackage[table]{xcolor}

\usepackage{array}

\usepackage{enumitem}
\usepackage{algpseudocode,algorithm}
\usepackage{bbm}
\usepackage{multirow}
\usepackage{xcolor}

\DeclareMathAlphabet\mathbfcal{OMS}{cmsy}{b}{n}
\newcommand{\mcc}{\mathbfcal{H}}
\newcommand\Hyp[3]{%
	\left(\begin{array}{@{\,}c@{\,}}
		#1 \\ #2 \end{array}\middle|\;#3 \right)}
\newcommand\dhp[2]{\mathbfcal{H}^{\alpha,\beta}_{#1}\left(#2;N\right)}
\newcommand\dhpd{\mathbfcal{H}^{\alpha,\beta}_{n}\left(x\right)}
\newcommand\dhpn[2]{\hat{\mathbfcal{H}}^{\alpha,\beta}_{#1}\left(#2;N\right)}
\newcommand\dhpnn[2]{\hat{\mathbfcal{H}}^{\alpha,\beta}_{#1}\left(#2\right)}

\newcolumntype{+}{!{\vrule width 2pt}}

\newlength\savedwidth

\usepackage{setspace} 
\renewcommand{\figurename}{Fig}

\makeatletter
\renewcommand{\@biblabel}[1]{\quad#1.}
\makeatother

\usepackage{lastpage,fancyhdr,graphicx}
\usepackage{epstopdf}
\fancyheadoffset[L]{2.25in}
\fancyfootoffset[L]{2.25in}
\begin{document}
\vspace*{0.2in}

\begin{flushleft}
{\Large
\textbf\newline{Fast Computation of Hahn Polynomials for High Order Moments} 
}
\newline
\\
Basheera M. Mahmmod\textsuperscript{1}, 
Sadiq H. Abdulhussain\textsuperscript{1}, 
Tom\'{a}\v{s} Suk\textsuperscript{2}, 
Abir Hussain\textsuperscript{3*}
\\
\bigskip
\textbf{1} Department of Computer Engineering, University of Baghdad, Baghdad, 10071, Iraq
\\
\textbf{2} Czech Academy of Sciences, Institute of Information Theory and Automation,\\ Pod vod\'{a}renskou v\v{e}\v{z}\'{\i} 4, 182\,08 Praha 8, Czech Republic
\\
\textbf{3} School of Computer Science and Mathematics, Liverpool John Moores University, Liverpool, United Kingdom
\\
\bigskip

%
%








\end{flushleft}
\section*{Abstract}
Discrete Hahn polynomials (DHPs) and their moments are considered to be one of the efficient orthogonal moments and they are applied in various scientific areas such as image processing and feature extraction. Commonly, DHPs are used as object representation; however, they suffer from the problem of numerical instability when the moment order becomes large. In this paper, an efficient method for computation of Hahn orthogonal basis is proposed and applied to high orders. This paper developed a new mathematical model for computing the initial value of the DHP and for different values of DHP parameters ($\alpha$ and $\beta$). In addition, the proposed method is composed of two recurrence algorithms with an adaptive threshold to stabilize the generation of the DHP coefficients. It is compared with state-of-the-art algorithms in terms of computational cost and the maximum size that can be correctly generated. The experimental results show that the proposed algorithm performs better in both parameters for wide ranges of parameter values of ($\alpha$ and $\beta$) and polynomial sizes.

\section{Introduction}
Moment theory and their variants are significant tools in imaging and computer vision applications \cite{TVS2020, 2D3D}.
Moments are scalar quantities used for characterization of the signal. 
Moments are obtained by using a set of polynomial basis functions. They are used to transform signal from time domain (such as speech) or spatial domain (such as image) into the transform domain \cite{Hameed2021,DKTT2017}.
Geometric moments and moments invariants were introduced by \cite{Hu1962} to deal with the problem of pattern recognition. 
 They are not orthogonal \cite{Zhu20071688_DD}, what causes numerical problems.

Continuous moments are obtained using continuous orthogonal polynomials (COPs), e.g. Zernike \cite{deng2016stable8H} or Chebyshev with transformed radius \cite{HOSNYcolor2019}.
Continuous moment functions are inaccurate because of two sources of errors: coordinate transformation of the image and the approximation of the continuous integral \cite{Mukundan2001}. As a result, during image reconstruction process, the image will be far from perfect due to discretization and approximation \cite{Mukundan2004_21H}. 

In order to surmount these shortcomings, researches oriented towards discrete orthogonal polynomials (DOPs). 
They show better capabilities in image reconstruction \cite{Mukundan2001, Mukundan2004_21H, Yap2003_}. Beside, discrete orthogonal moments (DOMs) have the ability to represent 1D and 2D signals without redundancy, high energy compaction, and spectral resolution properties \cite{Zhou2005_1B, SEA2019, den2021stable, Mahmmod2021SEA}. 
Several types of DOPs have been recently used for signal representation and feature extraction such as discrete Chebyshev polynomials \cite{Bello2018_6H, DTT2017}, discrete Krawtchouk moments \cite{Mahmmod2020, DKT_2021}, and discrete Hahn moments \cite{Yap2007_42H}. In addition, DOPs are used to solve linear functional differential equations \cite{mizel2008orthogonal}.

The robustness of DOPs is essentially based on some important properties such as energy compaction, efficient data processing, numerical stability, robust data analysis, extraction features from the signal, and localization \cite{DKTT2017,s21061999}. 
However, the remarkable properties of the most DOMs can be applied only to medium size images, they are not applicable for large size images or high moment orders \cite{Daoui2020_HH}.
This limitation is determined by the DOP overflow, fluctuation of the polynomial values, as well as the high computational cost. Thus, new recurrence algorithms for generating the higher orders are still developed, e.g. for Chebyshev \cite{DTT2017} and Krawtchouk \cite{Mahmmod2020} polynomials. Recently, researchers have discussed other DOPs such as Charlier polynomials \cite{CHP_2020} and Hahn polynomials \cite{Daoui2020_HH}. 

The recursive algorithms reduce the complexity of calculating the coefficients of DOPs and the propagation of errors \cite{Zhu2010_43H,SKTP2018}.
We can use either a single recursive formula with respect to degree $ n $ or a double recursive formula also with respect to spacial or time coordinate $ x $.
The problem of numerical instability is solved by calculating the DOP coefficients with respect to the variable $ n $. However, this calculation is not efficient, when the size of 1D or 2D signals becomes large. For instance, the coefficients of Chebyshev polynomials have numerical instabilities since the squared norm of the scaled Chebyshev polynomials assumes small values. Mukundan \cite{Mukundan2004_21H} propoesed the recurrence algorithm in the $ x $-direction to resolve this issue. After that, many researches began to work on this problem such as \cite{Zhu2010_43H}.

In general, an attention has been paid to the computation cost, which is considered to be an important point that subjects to ill-conditioning, therefore it is taking a substantial consideration in different researches \cite{spiliotis1996fast32H, Shu2010_30H}.
For Meixner moment coefficients, this drawback is resolved via fast and efficient calculation in \cite{Meixner2021}. For Chebyshev moments, a fast and stable method is proposed by Abdulhussain et al. \cite{DTT2017} for higher polynomial degree by a combination of the three-term recurrence relations in the $n$- and $x$-directions.
 Daoui et al. \cite{Daoui2020_HH} proposed a new method using a modified Gram-Schmidt orthogonalization process. 
	This method reduces the numerical error propagation during recursive computations. However, it is relatively slow.

Motivated by this problem for the discrete orthogonal Hahn moments, this study introduces a new algorithm to tackle this issue through composing two recurrence algorithms ($ n $- and $ x $-recurrence relations) and an adaptive threshold to stabilize the generation of the DHP coefficients. The present paper is organized as follows: in Section~\ref{sec:priliminaries}, the preliminaries and the existed three-term recurrence algorithms are presented. In Section~\ref{sec:proposed}, the proposed recurrence algorithm is presented. In Section~\ref{sec:experiments}, the experimental analysis is performed to evaluate the proposed recurrence algorithm. Finally, conclusions are drawn in Section~\ref{sec:conclusion}.

\section{Preliminaries}\label{sec:priliminaries}
The mathematical definitions and fundamentals of DHP and discrete Hahn moments (DHMs) are introduced in this section.

\subsection{The mathematical definition of DHP}
DHPs of the $ n $th degree $ \dhpd $ are defined as the solution of the difference relation \cite{Zhu2010_43H, Edge2017}, which is given by
\begin{equation}\label{Eq_1}
	\phi(x) \Delta \nabla \dhpd+\psi(x)\Delta \dhpd + \lambda_n \dhpd = 0 ,
\end{equation}

where $ \phi(x) $ and $ \psi(x) $ are first and second order functions, respectively, and $ \lambda_n $ is a constant. $ \Delta \dhpd $ and $ \nabla \dhpd $ represent the forward and backward difference, respectively. The values of $ \phi(x) $, $ \psi(x) $, and $ \lambda_n $ are given by \cite{Zhu2010_43H}
\begin{eqnarray}\label{Eq_2}
	\phi(x) &=& x(N+\alpha-x)\\
	\psi(x) &=& (\beta+1)(N-1)-(\alpha+\beta+2)x\\
	\lambda_n &=& n(\alpha+\beta+n+1) ,
\end{eqnarray}

where $ \alpha $ and $ \beta $ are the DHP parameters ($ \alpha $ and $ \beta > -1$ or also $ \alpha $ and $ \beta < -N$). The values of $ \Delta \dhpd $ and $ \nabla \dhpd $ are defined as follows
\begin{eqnarray}\label{Eq_3}
	\Delta \dhpd &=& \dhp{n}{x+1}-\dhp{n}{x} \label{Eq_3.1}\\
	\nabla \dhpd &=& \dhp{n}{x}-\dhp{n}{x-1} \label{Eq_3.2} .
\end{eqnarray}

From Eqs.~\eqref{Eq_3.1} and~\eqref{Eq_3.2}, $ \Delta \nabla \dhpd $ can be written
\begin{equation}\label{Eq_4}
	\Delta \nabla \dhpd = \dhp{n}{x+1}-2\dhpd+\dhp{n}{x-1} ,
\end{equation}

where $ n $ represent the polynomial degree, $ x $ is the signal index, and $ N $ is the polynomial size (the number of samples). The solution of the Eq.~\eqref{Eq_1} is
\begin{equation}\label{Eq_5}
	\dhp{n}{x}=\frac{(-1)^n \,(\beta+1)_n \,(N-n)_n}{n!}\, {}_3F_2\Hyp{-n,-x,n+1+\alpha+\beta}{\beta+1,1-N}{1} ,
\end{equation}
where $ {}_3F_2(\cdot) $ represents the generalized hypergeometric series which is given by
\begin{equation}\label{Eq_x}
	{}_3F_2\Hyp{a,b,c}{d,e}{z} = \sum\limits_{k=0}^{\infty} \frac{(a)_k \, (b)_k \, (c)_k}{(d)_k \, (e)_k \, k!} \left(z\right)^k
\end{equation}
and $ (\cdot)_k $ represents the rising factorial also known as Pochhammer symbol. It is given by
\begin{equation}\label{Eq_pochhamer}
	(a)_k = a(a+1)(a+2)\cdots(a+k-1)\ .
\end{equation}

DHPs satisfy the orthogonality condition\footnote{We can find alternative definition with different range of the orthogonality from 0 to $N$ \cite{Koekoek} instead of that from 0 to $N-1$ in \eqref{Eq_11}. Then, all other formulas are modified.} as follows
\begin{equation}\label{Eq_11}
	\sum\limits_{x=0}^{N-1}\dhp{n}{x} \dhp{m}{x} \omega_\mcc(x) = \rho_\mcc(n) \delta_{nm}\ ,
\end{equation}
where $ \delta_{nm} $ represents the Kronecker delta, $ \omega_\mcc $ and $ \rho_\mcc $ are the weight and norm functions of DHP
\begin{align}
	\omega_\mcc(x) &= \frac{\Gamma(N+\alpha-x)\Gamma(\beta+x+1)}{\Gamma(N-x)\Gamma(x+1)} \label{Eq_12w} \\
	\rho_\mcc(x) &= \frac{\Gamma(\alpha+n+1)\Gamma(\beta+n+1)(\alpha+\beta+n+1)_N}{(2n+\alpha+\beta+1)\Gamma(n+1)\Gamma(N-n)}. \label{Eq_12n}
\end{align}

The weighted DHP of $ n $th degree is given by
\begin{equation}\label{Eq_13}
	\dhpn{n}{x} = \dhp{n}{x}\sqrt{\frac{\omega_\mcc}{\rho_\mcc}} .
\end{equation}

\subsection{The definition of DHM}
DHMs are the signal (speech or images) projection on the DHP basis.
For two-dimensional signal (image), $ f(x,y) $, the DHMs, $ \eta_{nm} $, is computed as follows
\begin{align}
	\label{Eq_2D}
	&\eta_{nm} = \sum\limits_{x=0}^{N_1-1} \sum\limits_{y=0}^{N_2-1}\dhpnn{n}{x;N_1} \dhpnn{m}{y;N_2} f(x,y) \\
	&n = 0,1,\dots,N_1-1; \ \ \mathrm{and}\ \
	m = 0,1,\dots,N_2-1, \nonumber		
\end{align}
where $ N_1 \times N_2 $ is the size of the image $ f(x,y) $. The image can be reconstructed from Hahn moment domain into the spatial domain by
\begin{align}
	\label{Eq_2D_Rec}
	&\hat{f}(x,y)=\sum\limits_{n=0}^{N_1-1}\sum\limits_{m=0}^{N_2-1}\dhpnn{n}{x;N_1} \dhpnn{n}{y;N_2} \eta_{nm} \\
	&x = 0,1,\dots,N_1-1; \ \ \mathrm{and}\ \
	y = 0,1,\dots,N_2-1 \nonumber .
\end{align}

\subsection{Existing Recurrence Algorithms}
The three term recurrence relations are employed because of both the time consumption and the insufficient precision of the hypergeometric series in Eq.~\eqref{Eq_5}.
In this section, the existing recurrence algorithms and their analysis are briefly presented.

\subsubsection{The Three Term Recurrence Relation in the $ n $-direction (TTRRnd)}
The DHP of the $ n $th degree at the $ x $th index is given by \cite{Zhu2010_43H}
\begin{align}\label{Eq_nd}
	&\dhpn{n}{x} = \mathcal{\frac{AB}{E}} \, \dhpn{n-1}{x}+\mathcal{\frac{CD}{E}} \, \dhpn{n-2}{x} \\
	&\ \ n=2,3,\dots,N-1;\ \ \mathrm{and}\ \
	x=0,1,\dots,N-1 \nonumber ,
\end{align}
where the parameters of the $ n $-direction recurrence algorithms are
\begin{align}\label{Eq_nd_Param}
	\mathcal{A}&=x-\frac{\alpha-\beta+2N-2}{4}-\frac{(-\alpha^2+\beta^2)(\beta+\alpha+2N)}{4(\alpha+\beta+2n-2)(\alpha+\beta+2n)}  \nonumber\\
	\mathcal{B}&= \sqrt{\frac{n(\alpha+\beta+n)(\alpha+\beta+2n+1)}{(N-n)(\alpha+n)(\beta+n)(\alpha+\beta+2n-1)(\alpha+\beta+N+n)}} \nonumber\\
	\mathcal{C}&=-\frac{(\alpha+n-1)(\beta+n-1)(\alpha+\beta+N+n-1)(N-n+1)}{(\alpha+\beta+2n-2)(\alpha+\beta+2n-1)} \\
	\mathcal{D}&=\sqrt{\frac{n(n-1)(\alpha+\beta+n)(\alpha+\beta+n-1)(\alpha+\beta+2n+1)}{(\alpha+n)(\alpha+n-1)(\beta+n)(\beta+n-1)(N-n+1)(N-n)}} \times \nonumber \\
	&\, \, \, \, \, \, \, \, \sqrt{\frac{1}{(\alpha+\beta+2n-3)(\alpha+\beta+N+n)(\alpha+\beta+N+n-1)}} \nonumber\\
	\mathcal{E}&=\frac{n(\alpha+\beta+n)}{(\alpha+\beta+2n-1)(\alpha+\beta+2n)} \nonumber
\end{align}
with initial values
\begin{align}\label{Eq_nd_init}
	\dhpn{0}{x}&=\sqrt{\frac{\omega_\mcc(x)}{\rho_\mcc(0)}} \\
	\label{Eq_nd_init2}
	\dhpn{1}{x}&=\left[-(\beta+1)(N-1)+x(\alpha+\beta+2)\right]\sqrt{\frac{\omega_\mcc(x)}{\rho_\mcc(1)}} .
\end{align}

\begin{figure}
	\centering
	\begin{tabular}{@{}c@{}c}
		\multicolumn{2}{c}{
			\includegraphics[width=0.533\linewidth]{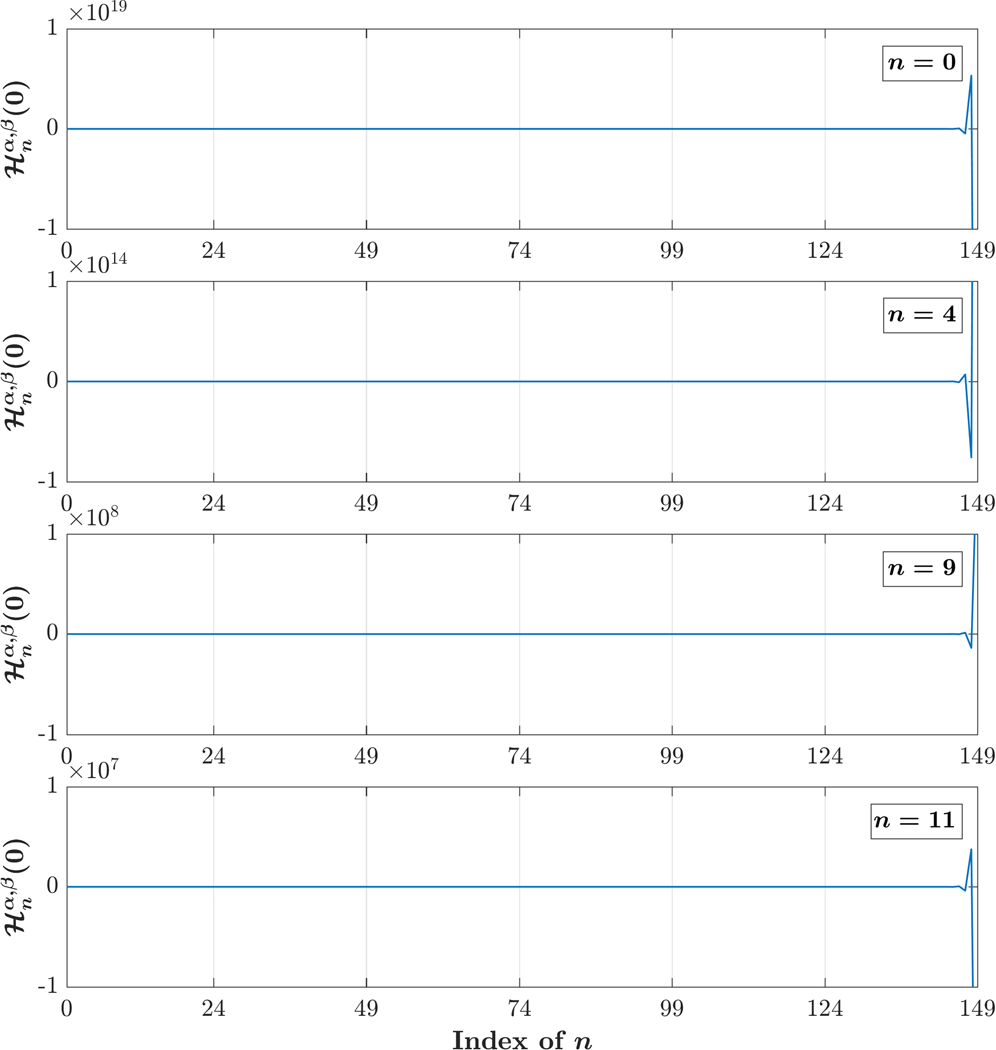}
		}
		\\
		\includegraphics[width=0.44\linewidth]{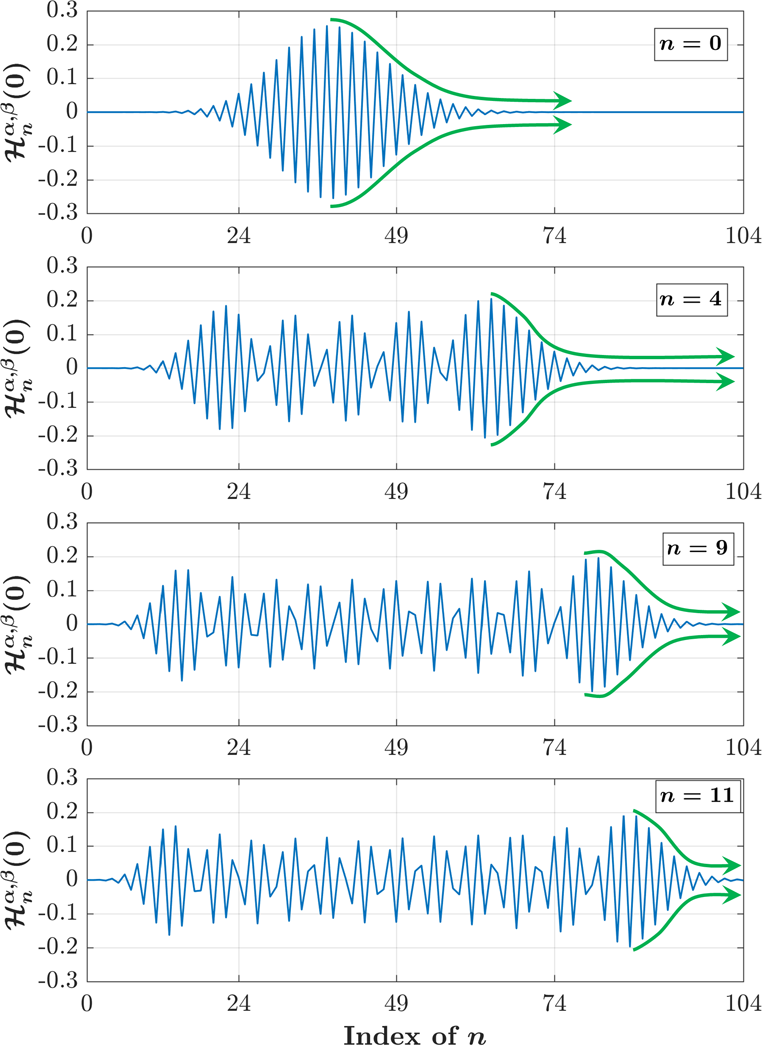}&
		\includegraphics[width=0.383\linewidth]{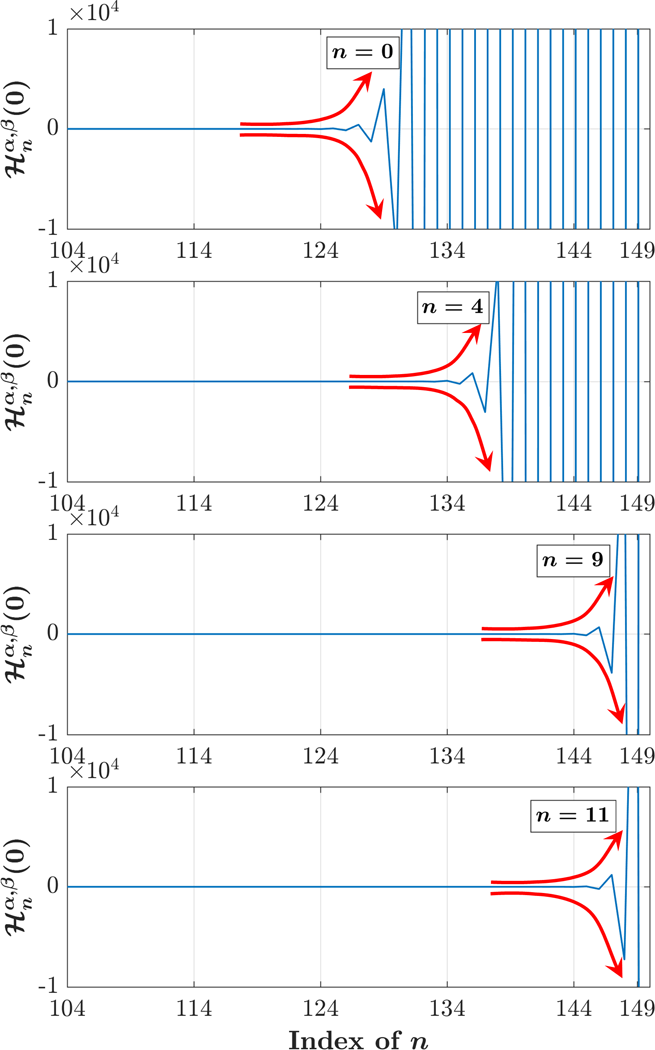}
	\end{tabular}
	\caption{Plot of DHP with different degree and a size of 150. Top row shows entire $ \dhpn{n}{0} $, bottom left shows stable values in the range of $ n=0,1,\dots,104 $, and bottom right column shows unstable values and the starting point of instability due to numerical errors in the range $ n=104,105,\dots,149 $.}
	\label{n-direction}
\end{figure}

The limitation of the $ n $-direction recurrence algorithm arises from the initial values $ \dhpn{0}{x} $ and $ \dhpn{1}{x} $.
They are bounded to limited polynomial size $ N $ and DHP parameters $ \alpha $ and $ \beta $. The maximum polynomial size that can be generated is 135 samples, when DHP parameters, $ \alpha $ and $ \beta $, are 20 and 20, respectively. The limitation arises from the nature of the formula used. Although this issue can be solved by reducing the complexity of the formula employed to compute the values of the initial sets, the $ n $-direction recurrence algorithm still suffers from the numerical propagation error. 
It is taken place when the values of the DHP coefficients (DHPCs) decrease in their values as shown in \figurename{~\ref{n-direction}}.

\subsubsection{The Three Term Recurrence Relation in the $ x $-direction (TTRRxd)}
DHP of $ n $th degree at the $ x $th index is computed as \cite{Zhu2010_43H}
\begin{align}\label{Eq_xd}
	&\dhpn{n}{x} =\eta_1\left[ \eta_2 \dhpn{n}{x-1} + \eta_3 \dhpn{n}{x-2} \right] \\
	&\hspace{3em}x=2,3,\dots,N-1;\ \  \mathrm{and} \ \
	n=0,1,\dots,N-1 \nonumber ,
\end{align}
where the coefficients of the $ x $-direction recurrence algorithm are
\begin{equation}
	\label{Eq_xd_param}
	\begin{array}{ll}
		\eta_1 = \displaystyle\frac{\sqrt{\omega_\mcc(x)}}{\sigma(x-1)+\tau(x-1)} & \sigma(x) = x(N+\alpha-x) \\
		\eta_2 = \displaystyle\frac{2\sigma(x-1)+\tau(x-1)-\lambda(n)}{\sqrt{\omega_\mcc(x-1)}} & \tau(x) = (\beta+1) (N-1) - x(\alpha+\beta+2) \\
		\eta_3 = \displaystyle - \frac{\sigma(x-1)}{\sqrt{\omega_\mcc(x-2)}} & \lambda(n) = n (\alpha + \beta + n +1)
	\end{array}
\end{equation}
with initial values
\begin{align}\label{Eq_xd_init}
	\dhpn{n}{0} & = (1-N)_n \, \binom{n+\beta}{n} \sqrt{\frac{\omega_\mcc(0)}{\rho_\mcc(n)}} \\ \label{Eq_xd_init2}
	\dhpn{n}{1} & = \frac{(n+\beta+1)(N-n-1)-n(N+\alpha-1)}{(\beta+1)(N-1)} \times \nonumber \\
	&\ \ \times
	\sqrt{\frac{\omega_\mcc(1)}{\ \omega_\mcc(0)}} \dhpn{n}{0}.
\end{align}

\begin{figure}[ht]
	\centering
	\includegraphics[width=0.99\linewidth]{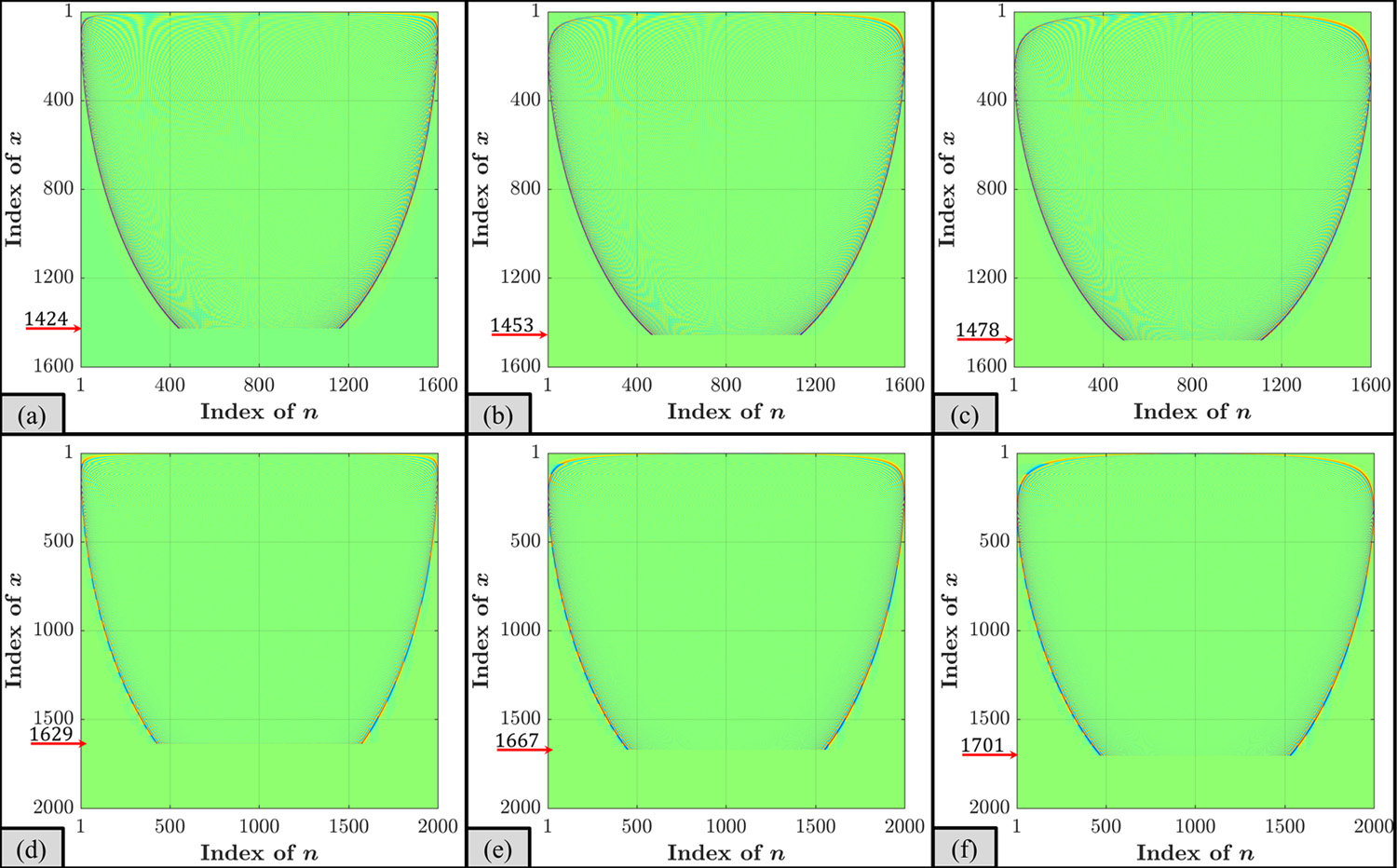}
	\caption{2D plots of the DHP. (a) $ N=1600 $ and $ \alpha=\beta=10 $, (b) $ N=1600 $ and $ \alpha=\beta=40 $, (c) $ N=1600 $ and $ \alpha=\beta=80 $, (d) $ N=2000 $ and $ \alpha=\beta=10 $, (e) $ N=2000 $ and $ \alpha=\beta=40 $, and (f) $ N=2000 $ and $ \alpha=\beta=80 $}
	\label{x-direction}
\end{figure}

To reduce the time required to compute the DHPCs, the following symmetry relation \cite{Daoui2020_HH} is employed
\begin{equation}\label{Eq_Sym}
	\dhpn{n}{x}=(-1)^n \dhpn{n}{N-1-x} \text{ for } \alpha=\beta .
\end{equation}
Using the symmetry relation, Eq.~\eqref{Eq_Sym} reduces the computed coefficients to 50\%. However, there are two limitations in the recurrence relation in the $ x $-directions as follows: 1) the initial set $ \dhpn{n}{0} $ becomes zero when the number of samples or the parameter values becomes too big because of the nature of the formula used in Eq.~\eqref{Eq_xd_init}, and 2) the coefficient values become underflowed as the degree of the polynomial becomes large; this is because of the initial values becomes less than $ 10^{-324} $, which becomes zero in various environments, such as Matlab and C++. \figurename{~\ref{x-direction}} shows DHP using different values of parameters $ \alpha $ and $ \beta $ as well as polynomial size using the recurrence relation in the $ x $-direction.

From \figurename{~\ref{x-direction}}, the DHPCs become zero as the polynomial degree increases. For instance, the maximum non-zero coefficients occurred at $ n=1423 $ when $ N=1600 $ and $ \alpha=\beta=10 $ (see \figurename{~\ref{x-direction}}a).

\subsubsection{Recurrence Relation Based on Gram-Schmidt orthonormalization process (RRGSOP)}
Recently, Daoui et al. \cite{Daoui2020_HH} presented an algorithm based on Gram-Schmidt orthonormalization process (GSOP) and the $ n $-direction recurrence relation to compute DHP. The GSOP is used to overcome the problem of the instability in the DHPCs. The presented algorithm begins with the computation of the initial sets $ \dhpn{0}{x} $ and $ \dhpn{1}{x}$. Then, the recurrence relation in the $ n $-direction is employed to compute the coefficients for $ n>1 $. Finally, the GSOP is applied to minimize the numerical errors generated by the $ n $-direction recurrence algorithm. However, the GSOP-based recurrence algorithm satisfies the orthogonality condition, it has three issues. First, the algorithm is not able to correctly generate the coefficients of the DHP when $ \alpha\ne\beta $, which is observed from the results in \cite{Daoui2020_HH}. Second, the algorithm is unable to generate DHP for a wide range of parameters $ \alpha $ and $ \beta $ because of the formula used to compute the initial values. Third, the GSOP-based recurrence algorithm has high computational cost due to the nested loops of the employed GSOP to minimize the error for each polynomial degree, which in turn increases the number of operations required to compute the coefficients of the DHP.

\section{Proposed Recurrence Algorithm}\label{sec:proposed}

\begin{figure}[ht]
	\centering
	\includegraphics[width=0.5\linewidth]{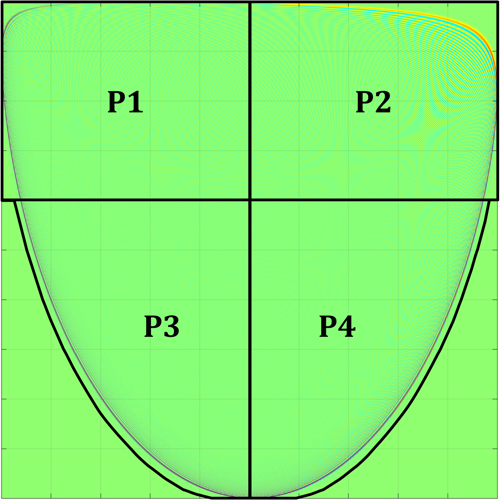}
	\caption{Schematic diagram of the proposed algorithm}
	\label{proposed}
\end{figure}

In this section, the design of the proposed algorithm is presented. For simplicity DHP $ \dhpn{n}{x} $ is denoted as $ \dhpnn{n}{x} $. \figurename{~\ref{proposed}} shows DHP for $ \alpha\ne\beta $, which is considered to be more general than the case $ \alpha=\beta $.

\subsection{Initial Value and Initial Sets}
The initial values are crucial and prerequisite for computation of the DHPC values. The existing algorithms (TTRRnd, TTRRxd, and RRGSOP) compute the initial sets. While TTRRnd and RRGSOP utilize Eqs.~\eqref{Eq_nd_init} and \eqref{Eq_nd_init2}, TTRRxd utilizes Eqs.~\eqref{Eq_xd_init} and \eqref{Eq_xd_init2} to determine the initial values. These equations are problematic because of the gamma and binomial functions. Thus, infinity (Inf) or not a number (NaN) are occurred in different environments such as C++, python, and MATLAB, i.e., the coefficients of the initial sets are not correctly computed. In addition, the existing algorithm computes the initial sets using the same formula, which in turn leads to increase the computation time.

From Eq.~\eqref{Eq_13}, the initial values are
\begin{align}
	\label{Eq_init_values1}
	\dhpnn{0}{0} &= \sqrt{\frac {\Gamma(\alpha+\beta+2) \Gamma(N+\alpha)}{\Gamma(\alpha+1) \Gamma(N+\beta+\alpha+1)}} \\
	\label{Eq_init_values2}
	\dhpnn{0}{N-1} &=\sqrt{\frac {\Gamma(\alpha+\beta+2) \Gamma(N+\beta)}{\Gamma(\beta+1) \Gamma(N+\alpha+\beta+1)}}.
\end{align}
They cannot be computed for a wide range of polynomial size and parameters, $ \alpha $ and $ \beta $, because of the gamma function $ \Gamma(\cdot) $ produces infinity for argument greater than 172 in the standard double precision arithmetics. 
The gamma function can be written as follows
\begin{equation}\label{Eq_logGamma}
	\Gamma(a)=e^{\mathrm{log}\Gamma(a)} ,
\end{equation}
where $ \mathrm{log}\Gamma(\cdot) $ represents the logarithmic gamma function \cite{Hart1978}.
Using Eq. \eqref{Eq_logGamma}, Eqs.~\eqref{Eq_init_values1} and \eqref{Eq_init_values2} can be expressed in terms of $ \mathrm{log}\Gamma(\cdot) $ functions
\begin{align}
	\label{Eq_init_val1}	
	\dhpnn{0}{0} &= e^{[\mathrm{log}\Gamma(\alpha+\beta+2)+\mathrm{log}\Gamma(N+\alpha)-\mathrm{log}\Gamma(\alpha+1)-\mathrm{log}\Gamma(N+\alpha+\beta+1)]/2} \\
	\label{Eq_init_val2}	
	\dhpnn{0}{N-1} &=  e^{[\mathrm{log}\Gamma(N+\beta)+\mathrm{log}\Gamma(\alpha+1)-\mathrm{log}\Gamma(\beta+1)-\mathrm{log}\Gamma(N+\alpha)]/2}\ \dhpnn{0}{0} .
\end{align}

	These formulas can be used for $\alpha >-1$ and $\beta >-1$. The parameters can also be less than $-N$, i.e. $\alpha <-N$ and $\beta <-N$. Then we must use slightly modified formulas
	\begin{align}
		\label{Eq_init_neg_val1}	
		\dhpnn{0}{0} &= e^{[-\mathrm{log}\Gamma(-\alpha-\beta-1)-\mathrm{log}\Gamma(-N-\alpha+1)+\mathrm{log}\Gamma(-\alpha)+\mathrm{log}\Gamma(-N-\alpha-\beta)]/2} \\
		\label{Eq_init_neg_val2}	
		\dhpnn{0}{N-1} &=  e^{[-\mathrm{log}\Gamma(-\alpha-\beta-1)-\mathrm{log}\Gamma(-N-\beta+1)+\mathrm{log}\Gamma(-\beta)+\mathrm{log}\Gamma(-N-\alpha-\beta)]/2}\ .
	\end{align}

These formulas are used for each part P1 and P2 in \figurename{~\ref{proposed}} separately.

\begin{figure}
	\centering
	\includegraphics[width=0.99\linewidth]{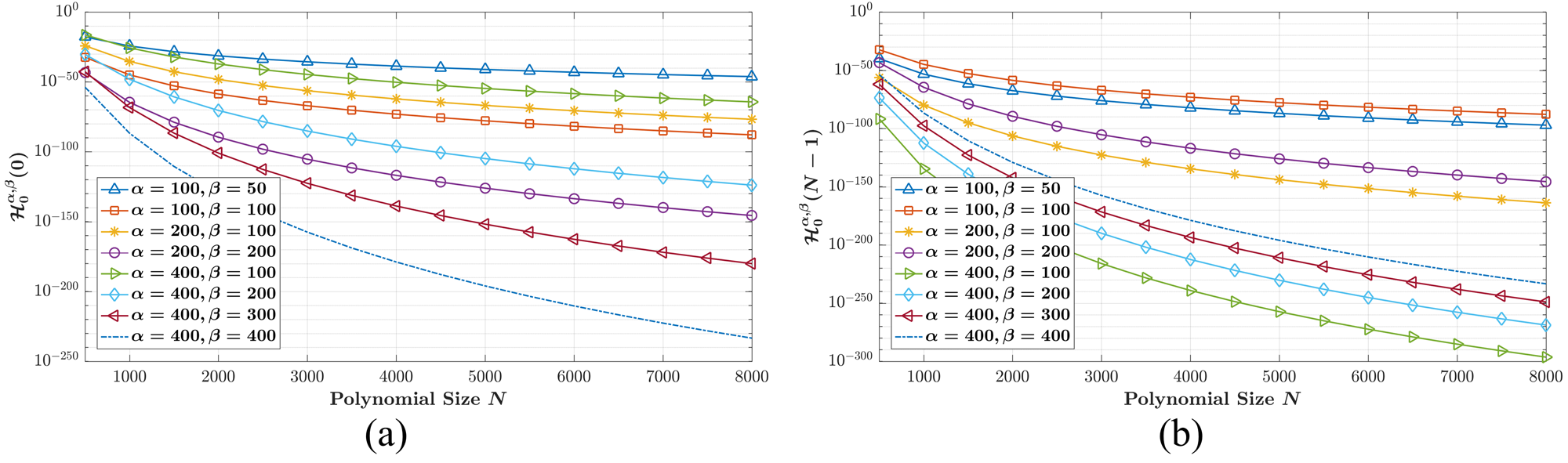}
	\caption{Plot of the initial using the proposed formula for $N=8000$ and different values of DHP parameters ($ \alpha $ and $ \beta $).}
	\label{initialsproposed}
\end{figure}

It is noteworthy that the initial value $ \dhpnn{0}{N-1} $ is computed in terms of $ \dhpnn{0}{0} $, as given in Eq.~\eqref{Eq_init_val2}, to reduce the execution time. \figurename{~\ref{initialsproposed}} illustrates the plots for the values of the initial values using the proposed algorithm for large polynomial size $ N=8000 $ and different values of DHP parameters $ \alpha $ and $ \beta $. From the \figurename{~\ref{initialsproposed}}, it can be observed that the initial values can be computed without Inf or NaN values; thus, the proposed initial formulas for the initial values can be used to compute the rest of the DHPCs.

It is well known that the initial sets are used with the aim of the three term recurrence algorithm for the polynomial computation. The initial sets for part P1 are $ \dhpnn{n}{0} $ and $ \dhpnn{n}{1} $ and for part P2 are $ \dhpnn{n}{N-1} $ and $ \dhpnn{n}{N-2} $. These sets are computed using two term recurrence algorithm. The initial sets for part P1 are computed as follows
\begin{equation}
	\label{Eq_initial_set_n_0}
	\begin{array}{l}
		\displaystyle
		\dhpnn{n+1}{0}\! =\! -\sqrt {\frac{(n+1+\beta)(N-1-n)(2n+3+\beta+\alpha)(n+\beta+\alpha+1)}{(n+1) (\alpha+1+n)(n+1+\beta+\alpha+N)(2n+\beta+\alpha+1)}} \times\\
		\hspace{6em}\times\dhpnn{n}{0},\hspace{6em}
		n=1,2,\dots,N-1
	\end{array}
\end{equation}
\begin{equation}
	\label{Eq_initial_set_n_1}
	\begin{array}{l}
		\displaystyle
		\dhpnn{n}{1}={\frac{(N-n-1)\beta-{n}^{2}-( \alpha+1)n+N-1}{(\beta+1)(N-1)}\sqrt {{\frac {( \beta+1)(N-1)}{\alpha+N-1}}}}\times\\
		\hspace{6em}\times \dhpnn{n}{0}, \hspace{6em}
		n=0,1,\dots,N-1 .
	\end{array}
\end{equation}		

The initial sets for part P2 are computed as follows
\begin{equation}
	\label{Eq_initial_set_n_N_1}
	\begin{array}{l}
		\displaystyle
		\hat{\mathbfcal{H}}^{\alpha,\beta}_{n+1}\!\left(N\!\!-\!\!1\right)
		\!=\!\!\sqrt {{\frac{(2n+3+\alpha+\beta)(n+\alpha+\beta+1)(n+1+\alpha)(N-n-1)}{(n+\beta+1)(n+1+\alpha+\beta+N)(2n+\alpha+\beta+1)(n+1) }}}\!\times\\
		\hspace{6em}\times \dhpnn{n}{N-1},\hspace{6em}
		n=1,2,\dots,N-1
	\end{array}
\end{equation}
\begin{equation}
	\label{Eq_initial_set_n_N_2}
	\begin{array}{l}
		\displaystyle
		\dhpnn{n}{N-2}={\frac {-{n}^{2}+( -\alpha-\beta-1)n+(N-1)(\alpha+1)}{\sqrt{(\beta+N-1)(N-1)(\alpha+1) }}} \dhpnn{n}{N-1},\\
		\hspace{5em}
		n=0,1,\dots,N-1 .
	\end{array}
\end{equation}

\subsection{Utilization of the TTRRnd and TTRRxd in the proposed algorithm}
The proposed recurrence algorithm is designed based on merging the TTRRnd and TTRRxd. For the parts P1 and P2, the coefficients of the DHP are computed using TTRRxd with a maximum degree of 40\% of $ N $. For the part P1, the modified TTRRxd (mTTRRxd) is applied as follows
\begin{align}
	\label{Eq_mTTRRxd}
	&\dhpnn{n}{x} = (\nu_1+\nu_2) \dhpnn{n}{x-1} + \nu_3 \dhpnn{n}{x-2} \\
	&\hspace{1.5em}n=0,1,\dots,M \ \ \mathrm{and}\ \
	x=2,3,\dots,N/2-1 , \nonumber
\end{align}

where $ M=0.4N $ to ensure non-zero initial sets. The parameters $\nu_1, \nu_2, $ and $ \nu_3$ of the mTTRRxd are defined by
\begin{align}
	\label{Eq_mxd_param}
	\nu_1&={\frac {-2\,{x}^{2}+ (2N+\alpha-\beta+2 ) x+(\beta-1)N-\alpha-1}{\nu}} \nonumber\\
	\nu_2&=-{\frac {n(\alpha+\beta+n+1) }{\nu}} \\
	\nu_3&=-{\frac{\sqrt{(\beta+x-1)(N-x+1)(x-1)(N+\alpha-x+1)}}{\nu}}\nonumber\\
	\nu&=\sqrt{(N-x)(\beta+x)(N+\alpha-x) x} .\nonumber
\end{align}

For the part P2, the mTTRRxd is applied backwardly as follows
\begin{align}
	\label{Eq_mTTRRxd_Back}
	&\dhpnn{n}{x-2} = \frac{1}{\nu_3} \dhpnn{n}{x} - \frac{\nu_1+\nu_2}{\nu_3} \dhpnn{n}{x-1} \\
	&n=0,1,\dots,M; \ \ \mathrm{and}\ \
	x=N-2,N-3,\dots,N/2 . \nonumber
\end{align}

After computing the coefficients in parts P1 and P2, the coefficients in parts P3 and P4 are computed using the modified TTRRnd (mTTRRnd) and mTTRRxd.

	The coefficients of part P3 are computed in the range $ n=M,M+1,\dots,N-1 $ and $ x=N/2-1, N/2-2, \ldots, 0 $ ($ x=(N-1)/2-1, (N-1)/2-2, \ldots, 0 $ for odd $N$) and the coefficients of part P4 are computed in the range $ n=M,M+1,\dots,N-1 $ and $ x=N/2, N/2+1, \ldots, N-1 $ ($ x=(N-1)/2, (N-1)/2+1, \ldots, N-1 $) as follows
	\begin{align}\label{Eq_mTTRnd}
		\dhpnn{n}{x}=\kappa_1\kappa_2 \dhpnn{n-1}{x} + \kappa_3 \dhpnn{n-2}{x} .
	\end{align}

The parameters $\kappa_1$, $\kappa_2, $ and $\kappa_3$ of the mTTRRxd are defined by
\begin{align}
	\label{eq_mTTRRnd_param}
	\kappa_1 =&(\alpha+\beta+2n) \sqrt{\frac{(\alpha+\beta+2n+1) (\alpha+\beta+2n-1)} {n(\alpha+\beta+n) (N-n) (\alpha+n) (\beta+n) (\alpha+\beta+N+n)}} \nonumber
	\\
	\kappa_2 =& x-\frac{1}{4}\left[\alpha-\beta+2N-2+{\frac{({\beta}^{2}-{\alpha}^{2})(\alpha+\beta+2N)}{(\alpha+\beta+2n-2)( \alpha+\beta+2n) }}\right]
	\\
		\kappa_3 =& -\frac{\alpha+\beta+2n} {\alpha+\beta+2n-2} \sqrt{\frac{(n-1) (\alpha+n-1) (\beta+n-1) (N-n+1)} {n(\alpha+\beta+n) (\alpha+n) (\beta+n)(N-n)}}
	\times \nonumber\\
	& \ \ \times\sqrt{\frac{(\alpha+\beta+n-1) (\alpha+\beta+2n+1) (\alpha+\beta+N+n-1)} {(\alpha+\beta+2n-3) (\alpha+\beta+N+n)}}
	 \nonumber
\end{align}

For each computed coefficient at the $ x $th index in parts P3 and P4, when the previously computed coefficient is less than the currently computed coefficient and less then $10^{-6}$, the mTTRRxd is stopped, the rest of the coefficients (part P6) is left zero and the computation continues by the next degree $ n+1 $.

\subsection{Summary of the Proposed Recurrence Algorithm}

\begin{figure} [ht]
	\centering
	\includegraphics[width=0.99\linewidth]{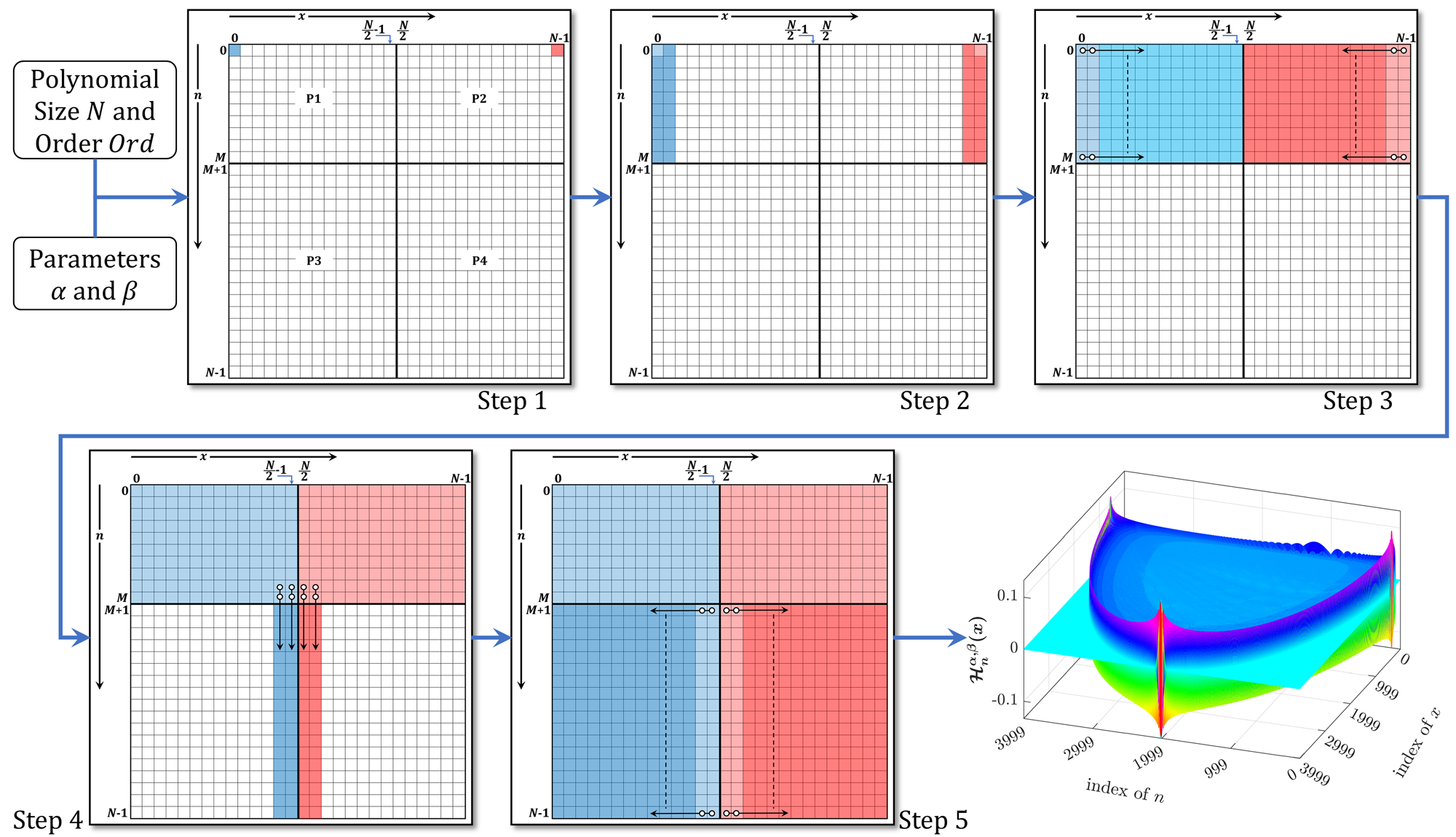}
	\caption{Steps of the proposed algorithm}
	\label{proposedsteps}
\end{figure}

For more clarification, the procedure of the proposed algorithm for $ \alpha\ne\beta $ are shown in \figurename{~\ref{proposedsteps}} and can be summarized as follows:

\begin{enumerate}[label=Step \arabic*:]
	\item Compute the initial values $ \dhpnn{0}{0} $ and $ \dhpnn{0}{N-1} $ using Eqs.~\eqref{Eq_init_val1} and \eqref{Eq_init_val2}, respectively. (Eqs.~\eqref{Eq_init_neg_val1} and \eqref{Eq_init_neg_val2} respectively).
	\item Compute the initial sets $ \dhpnn{n}{0} $ and $ \dhpnn{n}{1} $ using Eqs.~\eqref{Eq_initial_set_n_0} and \eqref{Eq_initial_set_n_1}. 
	The initial sets $ \dhpnn{n}{N-1} $ and $ \dhpnn{n}{N-2} $ are computed using Eqs.~\eqref{Eq_initial_set_n_N_1} and \eqref{Eq_initial_set_n_N_2}.
	\item Compute the coefficients of DHP in part P1 using mTTRxd given in Eq.~\eqref{Eq_mTTRRxd}, while the coefficients of DHP in part P2 are computed using backward mTTRRxd given in \eqref{Eq_mTTRRxd_Back}.
	\item Compute the coefficients of the DHP in parts P3 and P4 using backward mTTRxd given in \eqref{Eq_mTTRnd} for the range $ n=M+1,M+2,\dots,N-1 $. The coordinate $ x $ in P3 is in the range $ x = N2-1, N2-2, \dots, 0 $ and in P4 in the range $ x=N2, N2+1, \dots, N-1 $. $ N2 = N/2 $ for even $ N $ and $ N2 = (N-1)/2 $ for odd $ N $.
	\item For each value computed at the $ x $th index, if the absolute value of the previously computed coefficient is less than the currently computed coefficients and less than $10^{-6}$, the process is terminated and moved to the next value of $ n $, i.e., $ n+1 $. We call the zero coefficients part P6.
\end{enumerate}

It should be noted that for the case of $ \alpha=\beta $, the proposed algorithm computes the coefficients in parts P1 and P3, while the coefficients in part P5 are computed using symmetry relation given in Eq.~\eqref{Eq_Sym}.

\begin{figure}[ht]
	\centering
	\includegraphics[width=0.5\linewidth]{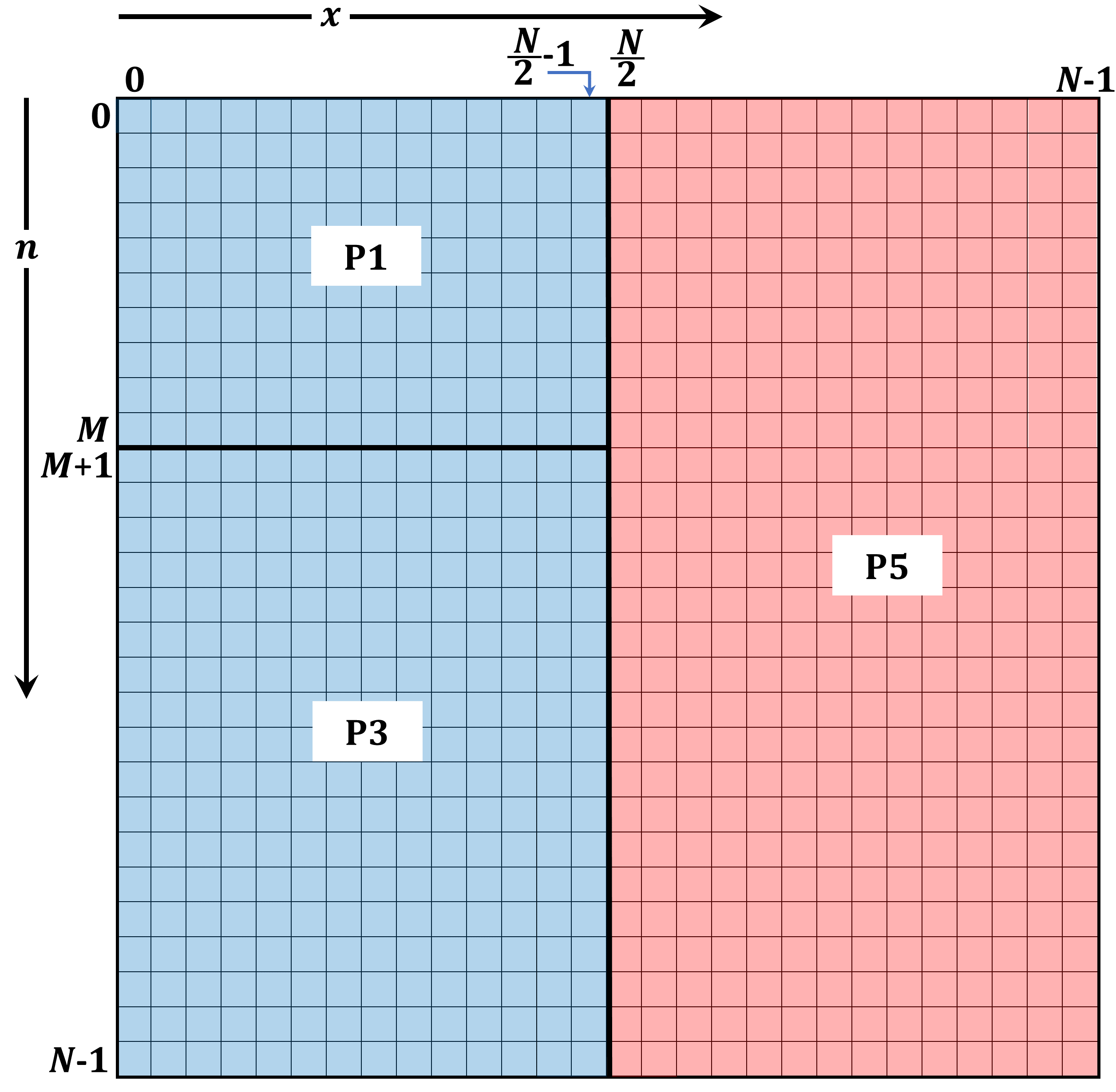}
	\caption{Part of the proposed algorithm for $ \alpha=\beta $}
	\label{proposed_ab}
\end{figure}

\section{Experimental Results}\label{sec:experiments}
The performance of the proposed algorithm is evaluated against existing algorithms in this section. The evaluation is performed in terms of the maximum size of the generated polynomial, signal reconstruction, and computational cost.

\subsection{Analysis of Maximum Size}


	In this experiment, we searched maximum size of DHP generated by the proposed and existing algorithms for different values of parameters $ \alpha $ and $ \beta $.
	If $ R $ is the matrix of values $ \dhpnn{n}{x} $ with the size $ N \times N $, then $ R \times R^T $ should be the identity matrix $\mathbbm{1}(N)$ of the size $N\times N$. So, $\mathrm{mean}(|\mathbbm{1}(N) - R \times R^T |)$ can be used as the average orthogonality error.
	We searched maximum $ N $ satisfying two criteria, average orthogonality error less than $10^{-5}$ and the computing time less than 1 minute.

\begin{table}[ht]
	\caption{Comparison between the ability of the proposed algorithm and existing algorithms to generate maximum polynomial size $N$ without propagation error.}
	\label{tbl_max_size}
	\vspace{1ex}
		\begin{center}
			\begin{tabular}{|c|c|c|c|c|}
				\hline
				$ \alpha   \text{ and } \beta $ & TTRRnd & TTRRxd & RRGSOP & Proposed \\ \hline
				$\alpha=100$; $ \beta=50$ & 35          & 1170 & $\emptyset$ & 9848  \\ \hline
				$\alpha=100$; $\beta=100$ & $\emptyset$ & 1309 & 1963        & 10749 \\ \hline
				$\alpha=200$; $\beta=100$ & $\emptyset$ & 1213 & $\emptyset$ & 10549 \\ \hline
				$\alpha=200$; $\beta=200$ & $\emptyset$ & 1414 & 1963        & 12037 \\ \hline
				$\alpha=400$; $\beta=200$ & $\emptyset$ & 1266 & $\emptyset$ & 11624 \\ \hline
				$\alpha=400$; $\beta=300$ & $\emptyset$ & 1369 & $\emptyset$ & 12907 \\ \hline
				$\alpha=400$; $\beta=400$ & $\emptyset$ & 1540 & 1970        & 14066 \\ \hline
				$\alpha=500$; $\beta=250$ & $\emptyset$ & 1285 & $\emptyset$ & 8747  \\ \hline
				$\alpha=500$; $\beta=400$ & $\emptyset$ & 1422 & $\emptyset$ & 11685 \\ \hline
				$\alpha=500$; $\beta=500$ & $\emptyset$ & 1589 & 1954        & 13527 \\ \hline
			\end{tabular}
		\end{center}
	
\end{table}

The maximum size for each algorithm is reported in \tablename{~\ref{tbl_max_size}}. From it we can observe that the proposed algorithm is able to generate DHP with different polynomial parameters without propagation error. The main problem with existing algorithms relies on: 1) the formula used to compute the initial value, and 2) the propagation error generated when the DHPCs becomes very small.

	The TTRRnd maximum size is 35 for a very small range of DHP parameters, while the TTRRxd is able to generate a maximum size of 1589 and it works for wide range of the DHP parameters. RRGSOP is able to generate the orthogonal polynomials only when $\alpha=\beta$. The main problem is the time of computation. All results $\neq\emptyset$ cannot be increased because of the time limit 1 minute, while the typical time of computation of the other algorithms is less than 2~s.
	The proposed algorithm outperforms the existing algorithms in terms of the maximum size and degree that can be generated.

\begin{sloppypar}
	\subsection{Analysis of the Energy Compaction and Reconstruction Error}
\end{sloppypar}

Discrete transforms are dissimilar because of their moments distribution \cite{CHP_2020}. The sequence of moment indices is essential for the reconstruction of signal information.
Therefore, the DHP distribution of the moment energy needs to be investigated before the signal reconstruction analysis. The procedure presented by Jian \cite{jain1989} has been followed to find the distribution of moments. The procedure can be summarized as follows
\begin{enumerate}
	\item A covariance matrix $ \mathcal{CM} $ with length $ N $ and zero mean is given by \cite{jain1989}
	\begin{equation}\label{Eq_cov_mat}
		\mathcal{CM}=\begin{bmatrix}
			1 		& 	\rho 	& 	\rho^2 	& 	\cdots 	& 	\rho^{N-1} \\
			\rho 	& 	1 		& 	\cdots 	& 	\cdots 	& 	\vdots 	\\
			\rho^2 	& 	\vdots 	& 	\ddots	& 	\vdots 	& 	\rho^2 	\\
			\vdots 	& 	\cdots 	& 	\cdots 	& 	\ddots 	& 	\rho \\
			\rho^{N-1} & 	\cdots 	& 	\rho^2 	& 	\rho 		& 	1 	
		\end{bmatrix} .
	\end{equation}
	
	\item The covariance matrix $ \mathcal{CM} $ is transformed into the discrete Hahn moment domain as follows
	\begin{equation}\label{Eq_CMM}
		T=R\times\mathcal{CM}\times R^T ,
	\end{equation}
	
	where $ T $ is the transformed matrix that is utilized to describe the transform coefficients $ \sigma_l^2 $, and $ R $ is the DHP matrix generated with a size of $ N $ and degree $ N $.
	
	\item The diagonal coefficients of the matrix $ T $ are considered
\end{enumerate}

The aforementioned procedure is carried out using two values of covariance coefficients, $ \rho=0.85 $ and $ \rho=0.95 $, different values of DHP parameters, and length $ N=16 $. The results are reported in \tablename{~\ref{tbl_cm}}.
It can inferred from \tablename{~\ref{tbl_cm}} that the variance values $ \sigma_l^2 $ of the DHP are arranged such that the maximum value is located at $ l=0 $ and the variance values reduced as the variance index increased. Thus, the DHP moment order, which is utilized to reconstruct signal information, is given by: $ n=0,1,\dots,N-1$.

\begin{table}[ht]
	\caption{Transform coefficient values $ \sigma_l^2 $}
	\label{tbl_cm}
	\centering
	\resizebox{\textwidth}{!}{
		\begin{tabular}{|c|c|c|c|c|c|c|c|c|c|c|c|c|}
			\hline
			\multirow{2}{*}{$l$} & \multicolumn{6}{c|}{$ \rho $=0.85} & \multicolumn{6}{c|}{$ \rho $=0.95} \\ \cline{2-13}
			& \begin{tabular}[c]{@{}c@{}}$\alpha=$20\\ $ \beta=$20\end{tabular} & \begin{tabular}[c]{@{}c@{}}$\alpha=$50\\ $ \beta=$50\end{tabular} & \begin{tabular}[c]{@{}c@{}}$\alpha=$100\\ $ \beta=$50\end{tabular} & \begin{tabular}[c]{@{}c@{}}$\alpha=$100\\ $ \beta=$100\end{tabular} & \begin{tabular}[c]{@{}c@{}}$\alpha=$200\\ $ \beta=$100\end{tabular} & \begin{tabular}[c]{@{}c@{}}$\alpha=$200\\ $ \beta=$200\end{tabular} & \begin{tabular}[c]{@{}c@{}}$\alpha=$20\\ $ \beta=$20\end{tabular} & \begin{tabular}[c]{@{}c@{}}$\alpha=$50\\ $ \beta=$50\end{tabular} & \begin{tabular}[c]{@{}c@{}}$\alpha=$100\\ $ \beta=$50\end{tabular} & \begin{tabular}[c]{@{}c@{}}$\alpha=$100\\ $ \beta=$100\end{tabular} & \begin{tabular}[c]{@{}c@{}}$\alpha=$200\\ $ \beta=$100\end{tabular} & \begin{tabular}[c]{@{}c@{}}$\alpha=$200\\ $ \beta=$200\end{tabular} \\ \hline
			0 & 9.145 & 8.635 & 8.031 & 8.437 & 7.896 & 8.331 & 6.729 & 6.458 & 6.121 & 6.350 & 6.046 & 6.292 \\ \hline
			1 & 2.713 & 2.850 & 2.672 & 2.886 & 2.679 & 2.902 & 2.622 & 2.434 & 2.214 & 2.359 & 2.157 & 2.318 \\ \hline
			2 & 1.336 & 1.255 & 1.260 & 1.332 & 1.291 & 1.372 & 2.228 & 2.267 & 2.140 & 2.274 & 2.135 & 2.276 \\ \hline
			3 & 1.053 & 1.215 & 1.141 & 1.167 & 1.100 & 1.142 & 1.287 & 1.333 & 1.291 & 1.343 & 1.282 & 1.347 \\ \hline
			4 & 0.676 & 0.705 & 0.779 & 0.711 & 0.768 & 0.713 & 0.986 & 1.104 & 1.128 & 1.148 & 1.140 & 1.170 \\ \hline
			5 & 0.346 & 0.472 & 0.603 & 0.530 & 0.638 & 0.563 & 0.586 & 0.673 & 0.780 & 0.708 & 0.793 & 0.727 \\ \hline
			6 & 0.290 & 0.346 & 0.527 & 0.369 & 0.538 & 0.380 & 0.409 & 0.494 & 0.633 & 0.533 & 0.656 & 0.555 \\ \hline
			7 & 0.107 & 0.139 & 0.315 & 0.162 & 0.339 & 0.176 & 0.253 & 0.299 & 0.453 & 0.323 & 0.478 & 0.337 \\ \hline
			8 & 0.098 & 0.136 & 0.259 & 0.151 & 0.289 & 0.160 & 0.183 & 0.211 & 0.338 & 0.227 & 0.363 & 0.237 \\ \hline
			9 & 0.047 & 0.054 & 0.144 & 0.058 & 0.165 & 0.061 & 0.138 & 0.148 & 0.237 & 0.154 & 0.258 & 0.159 \\ \hline
			10 & 0.041 & 0.047 & 0.093 & 0.051 & 0.108 & 0.054 & 0.117 & 0.120 & 0.170 & 0.123 & 0.185 & 0.124 \\ \hline
			11 & 0.033 & 0.034 & 0.055 & 0.034 & 0.062 & 0.034 & 0.105 & 0.104 & 0.128 & 0.105 & 0.135 & 0.105 \\ \hline
			12 & 0.031 & 0.031 & 0.038 & 0.031 & 0.041 & 0.031 & 0.097 & 0.096 & 0.105 & 0.096 & 0.108 & 0.096 \\ \hline
			13 & 0.029 & 0.029 & 0.031 & 0.029 & 0.031 & 0.028 & 0.091 & 0.090 & 0.093 & 0.090 & 0.094 & 0.090 \\ \hline
			14 & 0.027 & 0.027 & 0.028 & 0.027 & 0.028 & 0.027 & 0.087 & 0.086 & 0.087 & 0.086 & 0.087 & 0.086 \\ \hline
			15 & 0.026 & 0.026 & 0.026 & 0.026 & 0.026 & 0.026 & 0.083 & 0.083 & 0.083 & 0.083 & 0.083 & 0.083 \\ \hline
	\end{tabular}}
\end{table}

\begin{figure}[ht]
	\centering
	\includegraphics[width=0.90\linewidth]{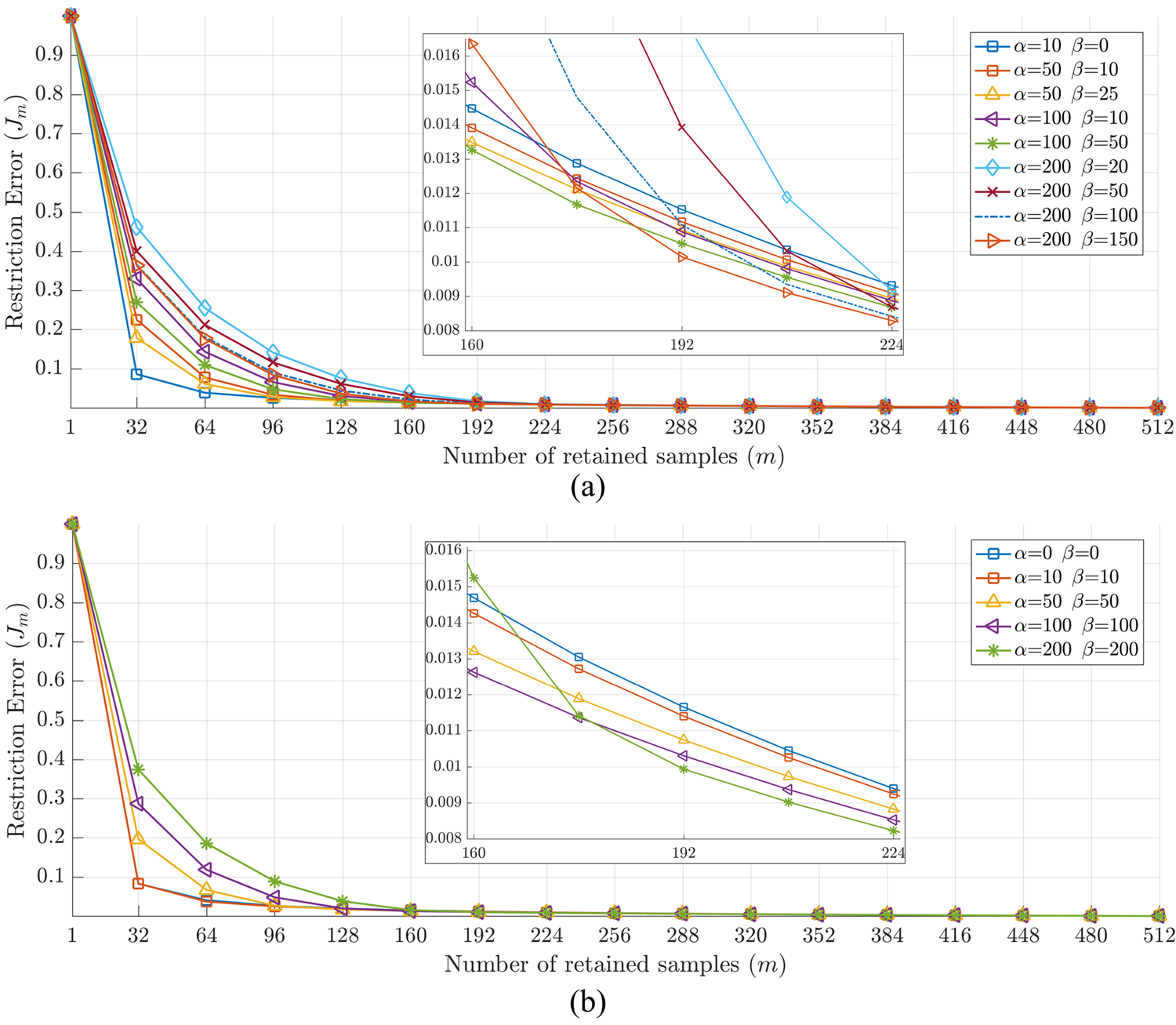}
	\caption{Restriction error of DHP (a) for $ \alpha\ne\beta $, and (b) $ \alpha = \beta $}
	\label{fig_jm}
\end{figure}

One of the important properties of the orthogonal-polynomial-based discrete transform is the energy compaction property. This property is employed to measure the tendency of the DHP to reconstruct a large amount of the signal information using a small number of moments coefficients. To examine the impact of the DHP parameters $ \alpha $ and $ \beta $ on the energy compaction, the restriction error, $ \mathcal{J} $, is used as follows \cite{jain1989}
\begin{equation}
	\label{Eq_Jm}
	\mathcal{J}_m=\frac{\sum\limits_{a=m}^{N-1}\sigma_a^2}{\sum\limits_{a=0}^{N-1}\sigma_a^2};\ \
	m=0,1,2,\dots,N-1 ,
\end{equation}
where $ \sigma_a^2 $ represents $ \sigma_l^2 $ ordered descendingly.
\figurename{~\ref{fig_jm}} illustrates the restriction error using covariance coefficient ($ \rho=0.98 $) in terms of retained samples~$ q $.
From \figurename{~\ref{fig_jm}}, the DHP parameters, $ \alpha $ and $ \beta $, affect the restriction error. E.g., from \figurename{~\ref{fig_jm}}a, when parameter $ \alpha=10 $ and $ \beta=0 $ shows better energy compaction than other parameter values in the range of $ m=32,\dots,96 $. However, when $ \alpha=200 $ and $ \beta=150 $ presents better energy compaction compared to other DHP parameters in the range $ m>192 $. On the other hand, \figurename{~\ref{fig_jm}}b, the energy compaction for DHP parameters $ \alpha=\beta=0 $ and  $ \alpha=\beta=10 $ shows comparable energy compaction as well as the best energy compaction in the range of retained samples $ m=32,\dots,96 $. It should be noted that as DHP parameters increases, the energy compaction becomes better in the range of $ m>192 $.

\begin{figure}[!ht]
	\centering
	\includegraphics[width=0.5\linewidth]{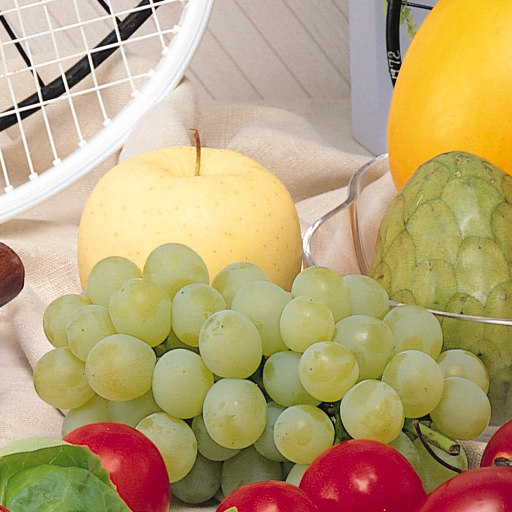}
	\caption{The image used for the tests.}
	\label{test_img}
\end{figure}

\begin{figure}[!ht]
	\centering
	\includegraphics[width=0.90\linewidth]{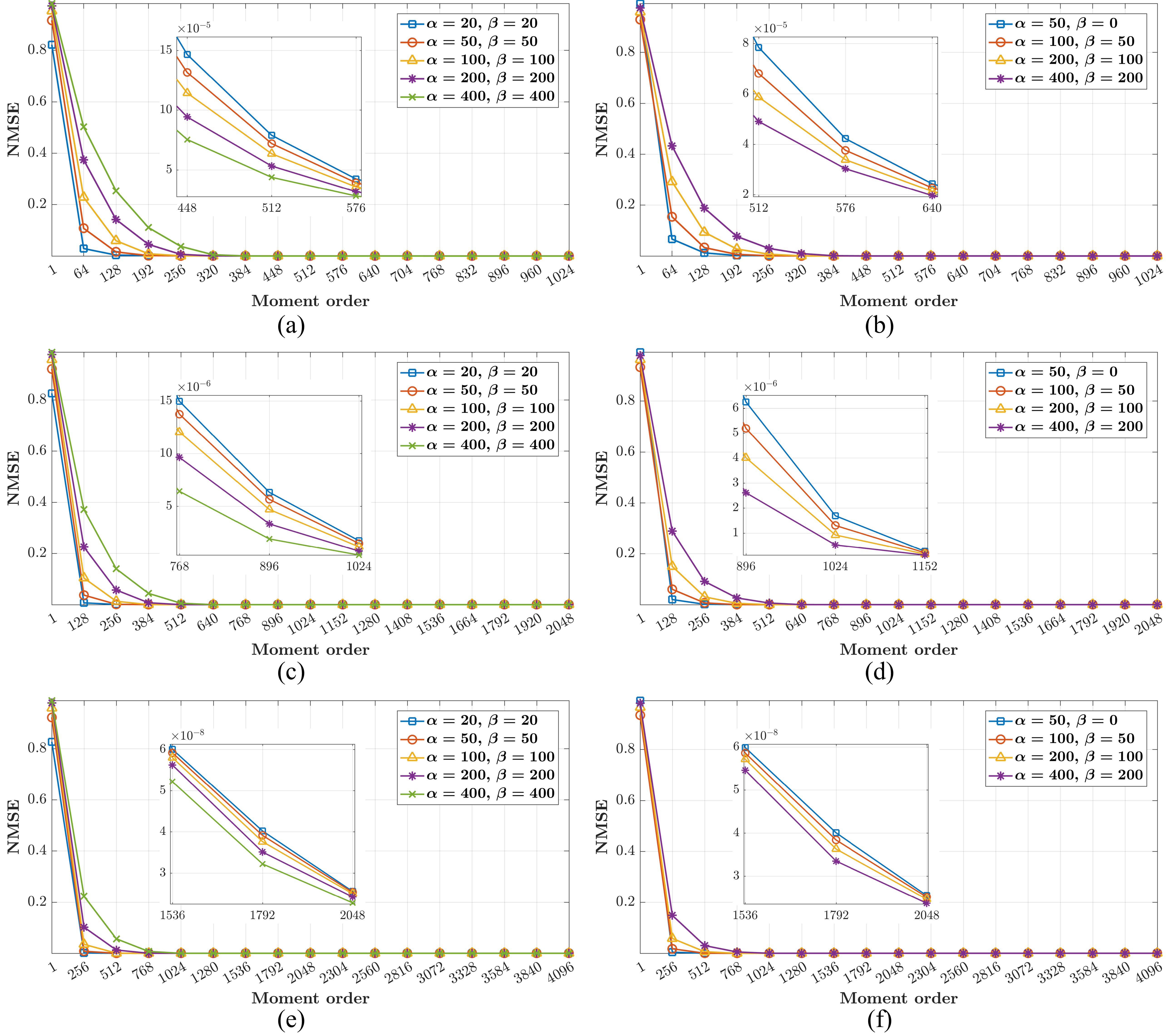}
	\caption{NMSE using the proposed algorithm for different values of parameters ($\alpha$ and $\beta$).}
	\label{fig_nmse}
\end{figure}

For more evaluation of the proposed algorithm, the normalized mean square error (NMSE) is computed between the image and its reconstructed version. The formula for NMSE, $ E $, is given in \cite{CHP_2020}
\begin{equation}
	\label{Eq_NMSE}
	E(I,I_r)=\frac{\sum\limits_{x,y}{(I(x,y)-I_r(x,y))^2}}{\sum\limits_{x,y}{(I(x,y))^2}}
\end{equation}

\begin{figure}[!ht]
	\centering
	\includegraphics[width=0.90\linewidth]{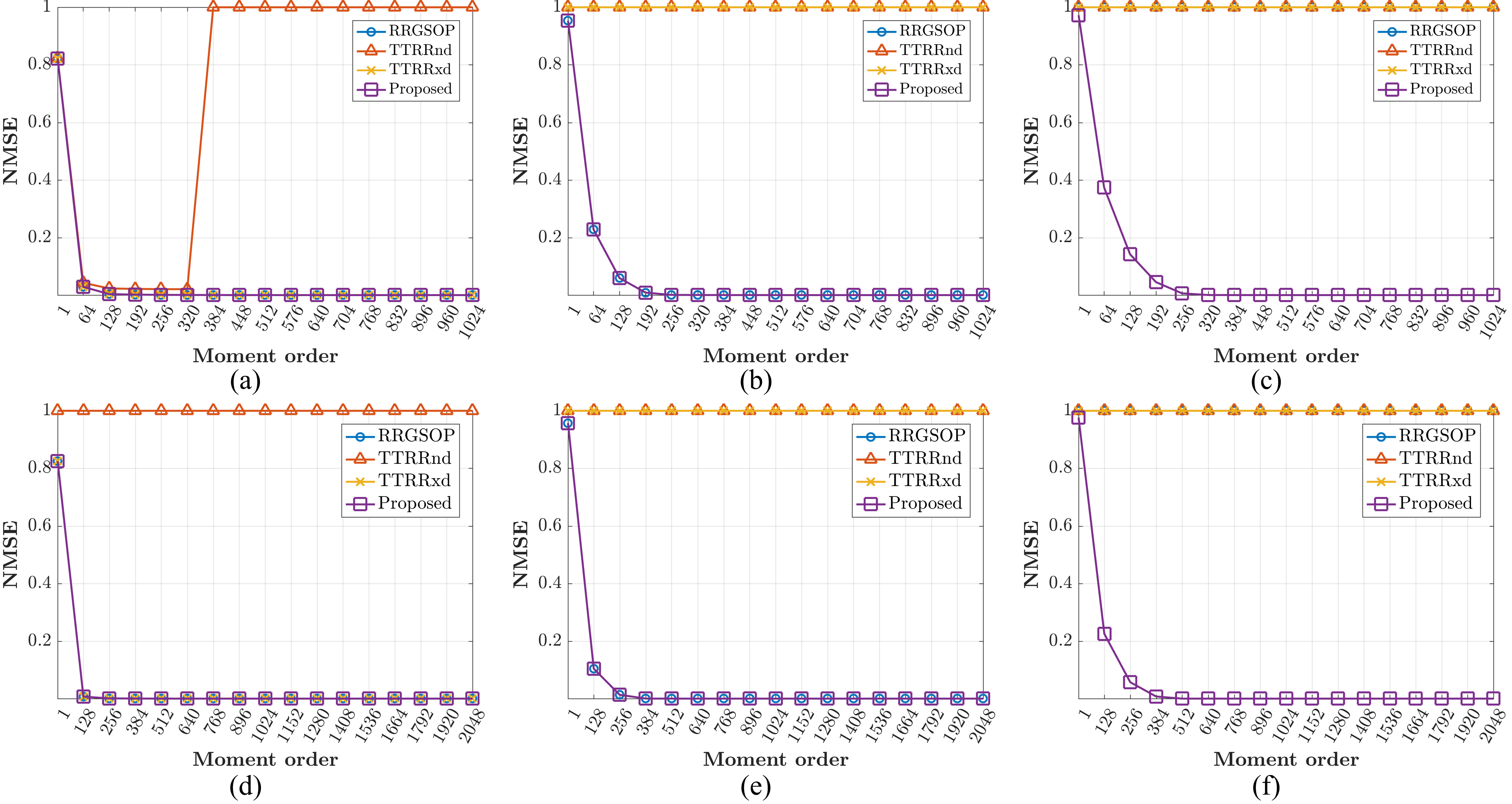}
	\caption{Comparison of NMSE between the proposed algorithm and existing works.}
	\label{fig_nmse_e}
\end{figure}

\begin{figure}[!ht]
	\centering
	\includegraphics[width=0.8\linewidth]{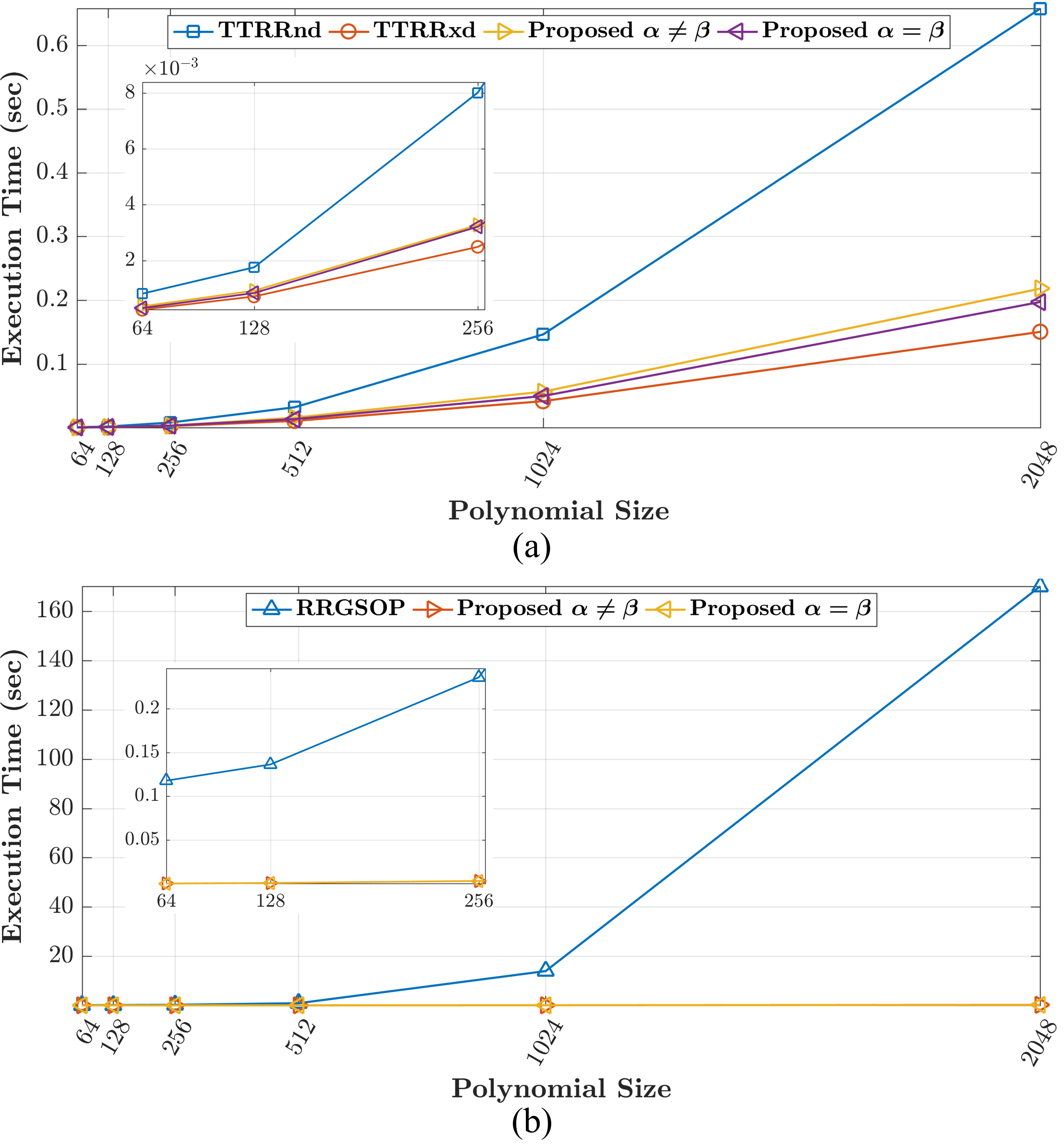}
	\caption{Comparison of execution time between the proposed algorithm and existing works.}
	\label{fig_time}
\end{figure}

The image "Fruits" shown in \figurename{~\ref{test_img}} is used as a test image for reconstruction error analysis. The reconstruction error is performed for different values of DHP parameters, $ \alpha $ and $ \beta $, as well as different polynomial size. For each polynomial size the image is resized accordingly.

First, the reconstruction error analysis is performed for the case of $ \alpha=\beta $ as shown in \figurename{s~}{\ref{fig_nmse}}a, c, and e and also for the case of $ \alpha\ne\beta $ as shown in \figurename{s~}{\ref{fig_nmse}}b, d, and f. The image sizes utilized are $ 1024\times1024 $, $ 2048\times2018 $, and $ 4096\times4096 $.

From \figurename{~\ref{fig_nmse}}, it can be concluded that the proposed algorithm is able to reconstruct the image remarkably for different polynomial size and different values of DHP parameters. It should be noted that when the DHP parameters increase to $ \alpha = \beta =400 $, it shows better reconstruction error than other values of DHP parameters. E.g., for image size $ 4096\times 4096 $, the reconstruction error, NMSE, is reduced to minimum for $ \alpha=\beta=400 $, when the number of moments used is greater than $ 1536\times1536 $.
In addition, for the case $ \alpha\ne\beta $, the best reconstruction error is occurred at $ \alpha=400 $ and $ \beta=200 $. For image size of $ 2048\times2048 $, the NMSE declined to minimum for DHP parameters $ \alpha=400 $ and $ \beta=200 $ before other DHP parameters when the moment matrix used for reconstruction is greater than $ 896\times896 $.

For comparison, the NMSE is computed using the proposed and existing algorithms for different values of DHP parameters and image sizes. \figurename{s~}{\ref{fig_nmse_e}}a, b, and c present the NMSE for image size of $ 1024\times1024 $ with DHP parameters $ \alpha=\beta=50 $, $ \alpha=\beta=100 $, and $ \alpha=\beta=200 $, respectively. In addition, \figurename{s~}{\ref{fig_nmse_e}}d, e, and f illustrate the NMSE for image size of $ 2048\times2048 $ with DHP parameters $ \alpha=\beta=50 $, $ \alpha=\beta=100 $, and $ \alpha=\beta=200 $, respectively.
It can be observed that the TTRRnd is unable to reconstruct the image because of the nature of the recurrence algorithm, where the coefficient value of DHP shows high propagation error for all values of DHP parameters, while for TTRRxd, it is able to reconstruct the image for small values of DHP parameters $ \alpha=\beta=50 $, as shown in \figurename{~\ref{fig_nmse_e}}a and \figurename{~\ref{fig_nmse_e}}d, and unable to reconstruct the image for large values of DHP parameters, as shown in \figurename{s~}{\ref{fig_nmse_e}}b, c, e, and f, where the TTRRxd is unable to generate correct DHPCs for large values of DHP parameters because of the formula used to compute the initial values. Whereas for RRGSOP, the algorithm is able to reconstruct the image correctly as the moment order increases up to $ \alpha=\beta=100 $; however, it is unable to reconstruct the image for DHP parameters greater than 100, as shown in \figurename{~\ref{fig_nmse_e}}c and \figurename{~\ref{fig_nmse_e}}d, because of the formula used to compute the initial values. On the other hand, the proposed algorithm is able to compute DHPCs and to reconstruct the image correctly for different values of DHP parameters and image sizes.

\subsection{Computational Cost Analysis}
In this section, the proposed algorithm is evaluated in terms of computational cost. The execution time is performed for the proposed algorithm and compared to that of the existing algorithms. The execution time experiment is carried out using different DHP sizes. The experiment is performed 10 times and the average time for each algorithm is reported as shows in  \figurename{~\ref{fig_time}}. It can be observed from \figurename{~\ref{fig_time}} that the execution time of the proposed algorithm for $ \alpha=\beta $ is less than that of the the proposed algorithm for $ \alpha\ne\beta $ in \figurename{~\ref{fig_time}}a. In addition, the proposed algorithm shows less computation time than that of the TTRRnd because of the proposed algorithm reduces the formula used for computation as depicted in \figurename{~\ref{fig_time}}a. Compared to TTRRxd, the proposed algorithm shows higher execution time than that of the TTRRxd. On the other hand, the execution time required to generate DHP using RRGSOP is higher than that of the proposed algorithm. The average improvement ratio computational cost for the proposed algorithm is $ \sim $0.76, 2.52, and 289.59 over TTRRxd, TTRRnd, and RRGSOP, respectively.

The experiment was carried out using MATLAB environment on MSI-GT60 laptop with a memory of 16GB and core i7-4700MQ CPU.

\section{Conclusion}\label{sec:conclusion}

	In this paper, a new recurrence algorithm for DHP is introduced. The proposed algorithm uses the logarithmic gamma function for computation of the initial values so that it is able to compute initial values for different DHP parameters and the large number of samples. The proposed algorithm combines recurrence in degree with recurrence in coordinates and with a condition criterion, so that the propagation error is suppressed.
	The experiments showed that the proposed algorithm remarkably reduces the computational cost with respect to the existing algorithms. It is able to eliminate propagation errors for large polynomial size and a wide range of DHP parameters. It achieves better signal reconstruction results than other recurrence algorithms for high polynomial degrees.

\section*{Acknowledgments}
\label{Acknowledgments}
This work has been supported by the Czech Science Foundation (Grant No. GA21-03921S) and by the {\it Praemium Academiae}. We would also like to acknowledge University of Baghdad for general support.

\nolinenumbers

%
%
%
%
%
%
%

\bibliographystyle{ieeetran}


\end{document}